\documentclass[10pt]{article}

\usepackage[letterpaper, left=1in, top=1in, right=1in, bottom=1in, verbose, ignoremp]{geometry}

\usepackage{url}\RequirePackage[colorlinks,citecolor=blue, linkcolor=blue,urlcolor = blue]{hyperref}
\usepackage{latexsym,amssymb,amsmath,amsfonts,graphicx,color,fancyvrb,amsthm,enumerate,subfigure,mathrsfs}
\usepackage[authoryear,round]{natbib}
\usepackage[dvipsnames]{xcolor}
\usepackage{xy}\xyoption{all} \xyoption{poly} \xyoption{knot}
\usepackage{float}
\usepackage{bm}
\usepackage{bbm}
\usepackage{multicol,multirow}
\usepackage{array}
\usepackage{relsize}
\usepackage{chngcntr}
\usepackage{etoolbox}
\usepackage{caption}
\usepackage{tikz}
\usepackage{tikz-cd}
\usetikzlibrary{patterns}
\usepackage{amsopn}
\usepackage{calc}
\usepackage{algorithm}
\usepackage[noend]{algpseudocode}
\usepackage{hyperref}

\usepackage{textcomp}

\newtheorem{theorem}{Theorem}[]
\newtheorem*{theorem*}{Theorem}

\newtheorem*{claim*}{Claim}
\theoremstyle{definition}
\newtheorem{definition}[theorem]{Definition}
\newtheorem*{definition*}{Definition}
\theoremstyle{AppDefinition}

\theoremstyle{AppClaim}

\theoremstyle{remark}

\newtheorem{example}[theorem]{Example}
\newtheorem*{example*}{Example}

\def\beginmat{ \left( \begin{array} }
\def\endmat{ \end{array} \right) }

\makeatletter
\newcommand*{\op}{%
  \DOTSB
  \mathop{\vphantom{\bigoplus}\mathpalette\matt@op\relax}%
  \slimits@
}
\newcommand\matt@op[2]{%
  \vcenter{\m@th\hbox{\resizebox{\widthof{$#1\bigoplus$}}{!}{$\boxplus$}}}%
}
\makeatother

\makeatletter
\def\@biblabel#1{}
\makeatother

\makeatletter
\patchcmd{\NAT@citex}
  {\@citea\NAT@hyper@{%
     \NAT@nmfmt{\NAT@nm}%
     \hyper@natlinkbreak{\NAT@aysep\NAT@spacechar}{\@citeb\@extra@b@citeb}%
     \NAT@date}}
  {\@citea\NAT@nmfmt{\NAT@nm}%
   \NAT@aysep\NAT@spacechar\NAT@hyper@{\NAT@date}}{}{}

\patchcmd{\NAT@citex}
  {\@citea\NAT@hyper@{%
     \NAT@nmfmt{\NAT@nm}%
     \hyper@natlinkbreak{\NAT@spacechar\NAT@@open\if*#1*\else#1\NAT@spacechar\fi}%
       {\@citeb\@extra@b@citeb}%
     \NAT@date}}
  {\@citea\NAT@nmfmt{\NAT@nm}%
   \NAT@spacechar\NAT@@open\if*#1*\else#1\NAT@spacechar\fi\NAT@hyper@{\NAT@date}}
  {}{}

\makeatother

\begin{document}
\def\spacingset#1{\renewcommand{\baselinestretch}%
{#1}\small\normalsize} \spacingset{1}
%\spacingset{1.45} % DON'T change the spacing!
\begin{flushleft}
{\Large{\textbf{Curved Markov Chain Monte Carlo for Network Learning}}}
\newline
\\
John Sigbeku$^{1}$, Emil Saucan$^{2}$, and Anthea Monod$^{1,\dagger}$
\\
\bigskip
\bf{1} Department of Mathematics, Imperial College London, UK 
\\
\bf{2} Department of Applied Mathematics, ORT Braude College of Engineering, Karmiel, Israel
\\
\bigskip
$\dagger$ Corresponding e-mail: a.monod@imperial.ac.uk
\end{flushleft}

%%%%%%%%%%%%%%%%%%%%%%%%%%%%%%%%%%%%%%%%%%%%%%%%%%%

\section*{Abstract}

We present a geometrically enhanced Markov chain Monte Carlo sampler for networks based on a discrete curvature measure defined on graphs.  Specifically, we incorporate the concept of graph Forman curvature into sampling procedures on both the nodes and edges of a network explicitly, via the transition probability of the Markov chain, as well as implicitly, via the target stationary distribution, which gives a novel, curved Markov chain Monte Carlo approach to learning networks.  We show that integrating curvature into the sampler results in faster convergence to a wide range of network statistics demonstrated on deterministic networks drawn from real-world data.

\paragraph{Keywords:} Discrete curvature; Graph Forman curvature; Markov chain Monte Carlo; Network learning; Network sampling

%%%%%%%%%%%%%%%%%%%%%%%%%%%%%%%%%%%%%%%%%%%%%%%%%%%

\section{Introduction}
\label{sec:intro}

The availability of abundant datasets present many new opportunities for insight into many fields of application, such as biology, social and telecommunications, and sensor analysis.   However, the computational burden of fully characterizing a network grows significantly with the size of the dataset and stands to hinder these potential insights.  A natural approach is to sample the larger networks to obtain a smaller representation for analysis, which drives the importance of obtaining an accurate representation by selecting the most pertinent characteristics of the network.  In this way, network sampling is an important means for learning networks.

Recent work incorporates the inherent geometry of the network in the sampling process, where the discrete curvature of the network has been shown to improve sampling efficiency and accuracy in a deterministic setting \citep{barkanass2020geometric}.  The idea is based on the observation that discrete curvature captures a notion of information flow on the network and thus is a natural concept to exploit when sampling networks with the aim to learn them.  Here, we adapt this approach to the random setting, which then makes statistical techniques of random sampling applicable to the problem of network learning.  Specifically, we present a curvature-based sampling approach based on the Markov chain Monte Carlo (MCMC) technique.  Our approach incorporates the {\em graph Forman curvature} of a network into the transition probabilities of a Markov chain defined on the nodes of a network as well as into the target stationary distribution.  In this way, our proposed methodology is both explicit and implicit.  We demonstrate our approach on deterministic networks arising from real-world data.  Results are bench-marked against the uniform Markov sampler to identify the cases where curvature-based sampling outperforms the base case.  We show that our geometrically enhanced MCMC sampler provides a quicker convergence to a wide set of network statistics.

The remainder of this paper is organized as follows.  We close this section with a brief overview of related work.  In Section \ref{sec:background}, we present the notion of graph Forman curvature; we also briefly overview the concept of MCMC sampling.  In Section \ref{sec:curved_mcmc}, we present our graph Forman-curved MCMC sampler for both edges and nodes.  We also give the network statistics that will serve as network characteristics to retain in the sampled subnetwork.  Section \ref{sec:results} presents details on the implementation and evaluation of our method; we demonstrate our approach on real-world deterministic networks.  We conclude with a discussion and opportunities for future research in Section \ref{sec:discussion}.

\paragraph{Related Work.}

Relevant to the ideas presented in this paper, edge- and node-based sampling methods, such as random edge and random node algorithms, draw from the statistical aspect of random sampling to sample representative networks \citep{leskovec2006sampling}.  Similarly, Markov chains on the nodes of the network with a specific stationary distribution have also been proposed, such as the Metropolis--Hastings random walk with uniform stationary distribution \citep{li2015random}.  

Existing sampling algorithms commonly assume full knowledge of the entire network, requiring its storage on local computer memory which can be computationally costly.  To bypass this expense, network crawling algorithms working on the basis of restricted access, where only the current node and its neighbors are required, have also been proposed, such as the forest fire sampling algorithm which visits a neighbor node with pre-specified probability \citep{leskovec2006sampling}.

%%%%%%%%%%%%%%%%%%%%%%%%%%%%%%%%%%%%%%%%%%%%%%%%%%%

\section{Overview: Discrete Curvature and Random Sampling}
\label{sec:background}

We now give a concise overview of the two main concepts which underlie the methodology we propose in this paper, namely, graph Forman curvature and Markov chain Monte Carlo.

\subsection{Discrete Curvature on Networks: Graph Forman Curvature}

Methods to handle data driven by geometric notions have recently inspired and established a variety of research domains such as topological data analysis, which aims to study the shape of data, and geometric deep learning, which incorporates geometric invariance into learning algorithms.  In computational settings, discretization is a fundamental consideration, giving rise to discrete geometry and, in particular, discrete curvature.  In this work, we study {\em graph Forman curvature}, which is the adaptation of the Forman--Ricci curvature to the edges of a network \citep{forman2003bochner,forman-networks,WJS16} .

For an edge $\langle i,j \rangle \in E$, graph Forman curvature takes into account its own weight, $w_{ij}$; the weights of the nodes it connects, $w(i)$ and $w(j)$; and the weights of incident edges excluding the edge itself $\langle i,j \rangle$, $e(i)\sim \langle i,j \rangle$ and $e(j)\sim \langle i,j\rangle$.

\begin{definition}
Given a network $G = \{V,E\}$, the {\em graph Forman curvature} $F(\cdot)$ for an edge $\langle i,j \rangle \in E$, is given by
\begin{equation}
\label{eq:forman_edge}
    F(\langle i,j \rangle) = w_{ij} \cdot \bigg( \frac{w(i)}{w_{ij}} + \frac{w(j)}{w_{ij}} - \sum_{e(i)\sim \langle i,j \rangle}\frac{w(i)}{\sqrt{w_{ij}w_{e(i)}}} -  \sum_{e(j)\sim \langle i,j \rangle}\frac{w(j)}{\sqrt{w_{ij}w_{e(j)}  }} \bigg).
\end{equation}
\end{definition}
Here, the edges in highly concentrated areas connecting nodes which are also connected to many other nearby nodes are given the highest absolute Forman curvature. 

In the combinatorial case, where each edge and node weight is set to one, (\ref{eq:forman_edge}) reduces to
\begin{equation}
\label{eq:forman_combinatorial}
F(\langle i,j \rangle) =  4 - d(i) -d(j),
\end{equation}
where $d(i)$ and $d(j)$ are the degrees of the nodes connected by $\langle i,j \rangle$. For further details, see \cite{SSWJ,forman-networks}.

\begin{figure}[ht]
    \centering
	\subfigure[]{
	\includegraphics[width=0.45\linewidth]{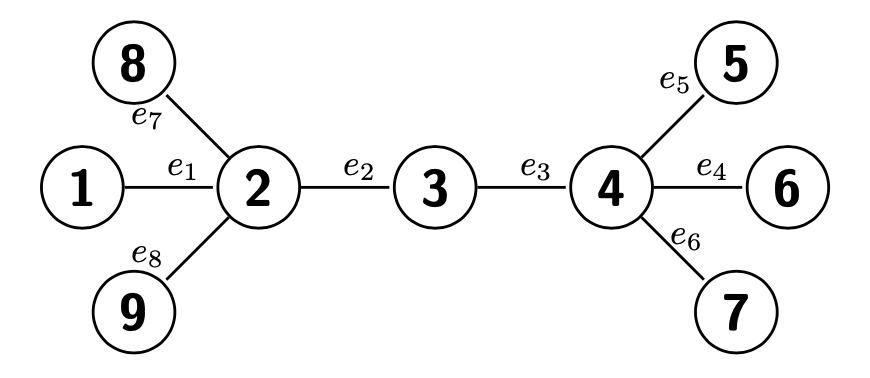}}
	\subfigure[]{
	\includegraphics[width=0.4\linewidth]{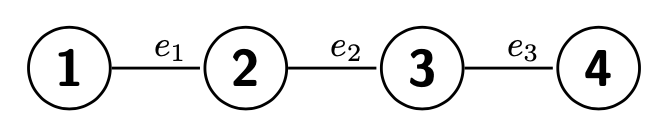}}
    \caption{Example networks to illustrate graph Forman curvature. (a) A combinatorial network; (b) A zero Forman curvature network.}
    \label{fig:ex_for}
\end{figure}

\begin{example}
Figure \ref{fig:ex_for}(a) illustrates how the graph Forman curvature captures the concentrated areas of a network. From (\ref{eq:forman_combinatorial}), we have $F(e_2)=F(e_3)= -2$, since both edges connect nodes with degrees two and four.  On the other hand, $F(e_1)=F(e_4)=F(e_5)=F(e_6)=F(e_7)=F(e_8)=-1$ which all connect nodes with degrees one and four. Edges $e_2$ and $e_3$ connect nodes with greater degrees than any of the remaining edges, and so they represent more concentrated areas of the network.  
\end{example}

\begin{example}
Figure \ref{fig:ex_for}(b) illustrates zero graph Forman curvature, assigned to any edge in a line-shaped combinatorial network where each connected node has degree of two, given by edge $e_2$ here.  This corresponds to the intuitive notion of curvature, where zero curvature corresponds to flatness.
\end{example}

Moreover, graph Forman curvature can be extended onto the nodes of network by summing the individual curvatures of the edges incident to a node. For a node $i$, this can be written as 
\begin{equation}
\label{eq:forman_node}
F(i)= \sum_{e(i)} F(\langle i,k \rangle), 
\end{equation}
where $e(i)$ are the edges incident to node $i$.  In our work, we apply both node- and edge-based Forman curvature to our sampling methodology.

\subsection{Random Sampling: Markov Chain Monte Carlo}

Probability-based inference entails estimating the expected value of a statistic or density from a probabilistic model, which can often be intractable in complex and high-dimensional settings.  Approximation procedures have been developed to overcome these difficulties; one such procedure is Markov chain Monte Carlo (MCMC).

In the network setting where there is inherent dependence, the statistical sampling procedure applied must necessarily take the dependence structure into account.  MCMC is a well-established technique comprising a class of algorithms to draw samples from high-dimensional and complex distributions that allows for a dependence structure in sampling, where the next sample drawn takes into account the existing sample; see e.g., \cite{murphy2012machine}.

Briefly, Monte Carlo allows for random sampling from a probability distribution to approximate a quantity or statistic of interest.  These methods typically sample independently and assume efficiency of sampling from the desired {\em target} or {\em stationary distribution}.  A Markov chain is a sequence of random variables $X_1, X_2, \ldots$ where each random variable in the sequence depends only on the variable drawn prior.  Given this dependence, there is then a specified {\em transition probability} between each sequential pair of random variables,  
$$
\mathbb{P}(X_{n+1} = x \mid X_1 = x_1,\, X_2 = x_2, \ldots,\, X_n = x_n) = \mathbb{P}(X_{n+1} = x \mid X_n = x_n).
$$
Merging the two concepts gives rise to the technique of MCMC, which is used to perform inference for probability distributions where independent samples cannot be drawn.  

In this setting, Monte Carlo cannot be directly applied, so samples are drawn from the probability distribution by constructing a Markov chain.  With enough iterations, the chain will eventually settle, or find equilibrium, on our quantity of interest (target distribution).  Since the purpose of the sample is still to approximate a quantity of interest, it technically is a Monte Carlo sample.  Thus, MCMC sampling generates a Monte Carlo sample that is drawn from Markov chain, which incorporates probabilistic dependence. 

%%%%%%%%%%%%%%%%%%%%%%%%%%%%%%%%%%%%%%%%%%%%%%%%%%%

\section{Forman-Curved Markov Chain Monte Carlo}
\label{sec:curved_mcmc}

For a given network $G = \{V,E\} $, we are interested in obtaining a sample subnetwork $G_n = \{V',E'\}$ with $V' \subseteq V$ and $E' \subseteq E$.  We would like for this subnetwork to capture pertinent characteristics of the larger network in order to serve as a faithful representation.  In the context of MCMC, we are interested in estimating these characteristics, which take the form of {\em network statistics}, by sampling.  The intuition behind incorporating the geometry of the network is that the additional information will improve the sampling method, either by obtaining more accurate estimates or quicker convergence to the target value of interest, i.e., the network statistics.  In particular, we incorporate graph Forman curvature into the sampling procedure, so that the sequence of random variables in the Markov chain are driven by curvature of the network.  This idea can be considered intuitively as regions of high curvature driving the direction of Markov chain.

In this section, we propose two methods of incorporating graph Forman curvature into network sampling: an explicit {\em edge-based} approach, and an implicit {\em node-based} approach.  We also give the network statistics that our proposed sampling method aims to approximate.

\subsection{The Edge-Based Approach}
\label{sec:edgeMCMC}

The edge-based Forman-curved MCMC sampler that we now propose explicitly integrates curvature into the transition probability of a Markov chain.

Using MCMC, $G_n$ may be obtained from the Markov chain on the set of nodes $V$ that is the sequence of random variables $X_1,X_2,\ldots,X_n$, where the actualizations are nodes, and the transition probability between nodes is given by
\begin{align}
\label{eq:mcmc_edge}
p_{ij} & = \mathbb{P}(X_{k+1}= j \mid X_1 = i_1 , X_2 = i_2,\ldots,X_k = i) = \mathbb{P}(X_{k+1}= j \mid X_k =i) \nonumber\\
& \propto \frac{|F(\langle i,j \rangle)|}{d(j)}
\end{align}
for all $i,j \in V$, and where $F(\langle i,j \rangle)$ is given by (\ref{eq:forman_edge}).

%In this work we study the performance of the edge-based Forman sampler (3) against the edge-based Uniform baseline (4):

%   \begin{align}
%    p_{ij}  \propto \frac{|F(\langle i,j \rangle)|}{d(j)} \\
%    p_{ij} = \frac{1}{d(j)}
%    \end{align}

Notice that this Markov property holds for all $k$, making this Markov chain time-homogeneous.  Also, two graph samples $G_n$ and $G_{n+1}$ are identical if the node $X_{n+1}$ has already been sampled, even though their Markov chains differ in length.  The chain can be initialized at different starting nodes $X_0$ to obtain different chains and graph samples.

\subsection{The Node-Based Approach}
\label{sec:nodeMCMC}

We now shift the focus from integrating curvature to alter transition probabilities of the Markov chain which generates node samples, to generating samples from a target probability distribution defined on the nodes of the network.  This approach implicitly incorporates curvature by defining a target distribution based on the graph Forman curvature around the nodes of the network.  In particular, a Forman-curved distribution where nodes are sampled with probability
\begin{equation}
\label{eq:mcmc_node}
\mathbb{P}(X_{k}= i) \propto \frac{|F(i)|}{d(i)}
\end{equation}
for large enough $k$, where $F(i)$ is given by (\ref{eq:forman_node}).  %In contrast, the Uniform target distribution is given by $\mathbb{P}(X_{k}= i) = {1}/{|V|}$
This target distribution can be thought of as our desired stable state, where in the long run as we continue to sample, we tend towards drawing node $i$ $100 \cdot \mathbb{P}(X_{k}= i)$ times per hundred draws. 

The Metropolis--Hastings (MH) algorithm is an MCMC algorithm often used in Bayesian statistics to generate samples from a specified target distribution up to a constant of proportionality; it is useful when the normalizing factor for a distribution is unknown.  It is implemented by sampling a known proposal probability distribution $q$ and applying a corrective procedure.  In our setting, we have the additional benefit of employing an MH algorithm to generate samples from the graph Forman-curved target distribution by crawling the network.  

\begin{algorithm}
	\caption{Metropolis--Hastings for Networks}
	\label{alg:MH}
	\begin{algorithmic}[1]
		\State $\textbf{input}$ Unnormalized density $g$; proposal distribution $q$; starting node $X_0$
		\State $\textbf{for} \ k=1,2,\ldots \textbf{do}$
		\State Set $Y_k \sim q(X_{k-1})$
		\State Set $U\sim \mathrm{Unif}(0,1)$
		\If {$\displaystyle U \leq \min\bigg(\frac{g(Y_k)q(Y_k,X_{k-1})}{g(X_{k-1})q(X_{k-1},Y_k)},\, 1 \bigg)$} 
		\State $X_k=Y_k$
		\Else
		\State $X_k=X_{k-1}$
		\EndIf
	\end{algorithmic}
\end{algorithm}

In our implementation of Algorithm \ref{alg:MH}, $g$ is the target distribution with $g(i)= |F(i)|/d(i)$ and the proposal distribution is given by $q(X_{k-1},Y_k) = 1/ d(i)$ when $X_{k-1} = i$ so that each neighbor node is proposed with equal probability.

In this node-based approach, the graph Forman curvature can be thought to implicitly influence the transition probability of the Markov chain, since the algorithm implies
$$
p_{ij} = \begin{cases}
            \displaystyle q_{ij} \cdot \frac{g(j)q_{ji}}{g(i)q_{ij}} & \text{ if } \displaystyle \frac{g(j)q_{ji}}{g(i)q_{ij}} \leq 1; \\
            q_{ij} & \text{ otherwise.}
        \end{cases}
$$

In contrast to the edge-based approach, this approach requires full access of the network to first calculate the target distribution. This is because the full set of network weights are required to calculate Forman curvature around all nodes in the network (i.e., $\{F(1),F(2),...,F(|V|) \}$). If the target distribution is not calculated beforehand, full access will be required once the Markov chain reaches a node that is connected to all other remaining nodes in the network, since evaluating the transition probabilities will require the set $\{F(1),F(2),...,F(|V|) \}$.\\

In both approaches, the degree is used as the normalization factor for the transition probabilities in order to bias sampling towards nodes with larger Forman curvature on average, in the combinatorial sense.  

%However, normalizing by strength, which results in bias towards nodes with largest weighted average, is a natural extension.

\subsection{Network Statistics}
\label{sec:stats}

Network statistics capture characteristics of the network in order to assess the faithfulness in representation of the sampled network in reference to the original larger network.  In this work, we focus on {\em centrality} network statistics, which describe different notions of node importance.  We apply our proposed Forman-curved MCMC samplers given above in Sections \ref{sec:edgeMCMC} and \ref{sec:nodeMCMC} to estimate the following network statistics.  We specify a wide variety of network statistics in order to study how well our sampling procedure is able to capture different characteristics of the larger network, e.g., \cite{airoldi2011network}.

%The following definitions rely on the shortest path as the minimum distance between any pair of nodes in a network.  Specifically, let $(v_1, v_2, \ldots ,v_n) \in V \times \cdots \times V$ be a sequence of adjacent nodes so that $v_i$ is adjacent to $v_{i+1}$ for $ 1 \leq i \leq n-1$, then this sequence is a path of length $n-1$ from node $v_1$ to $v_n$.  Let $(e_1, e_2,...,e_{n-1}) \in E \times \cdots \times E$ be the corresponding set of edges connecting the path, i.e., $e_i = (v_{i}, v_{i+1})$.  Then the shortest path $P(\cdot, \cdot)$ from $v_1 = i$ to $v_n = j$ is
%$\displaystyle P(v_1, v_n)= \underset{(v_1, v_2,\ldots,v_n)}{\operatorname{argmin}} \sum_{i=1}^{n-1} e_i.$
%In other words, it is the path, over all possible paths from $i$ to $j$, with the smallest distance.

\begin{definition}
\label{def:BCCC}
For a given network, let $\sigma_{i j}$ be the number of shortest paths $P(\cdot, \cdot)$ from node $i$ to $j$ and $\sigma_{i j}(i')$ be the number of those shortest paths passing through node $i'$.  The {\em betweenness centrality} $BC(\cdot)$ of a node $i'$ is given by
\begin{equation*}
\label{eq:BC}
    BC(i') = \sum_{i \neq i^{'} \neq j} \frac{\sigma_{i j}(i')}{\sigma_{i j}}.
\end{equation*}

The {\em closeness centrality} $CC(\cdot)$ of a node $i$ is given by
%\begin{equation}
%\label{eq:CC}
$\displaystyle
    CC(i) = \frac{1}{\sum_{j\neq i}P(i, j)}.
$
%\end{equation}
\end{definition}
Intuitively, betweenness centrality captures how much a node acts as a bridge between groups of nodes in a network; it captures how important a node is in connecting separate groups of nodes together.  Closeness centrality measures how close one node is to any other node in the network.

\begin{definition}
\label{def:strength}
Let $W = (W_{ij}) \in \mathbb{R}^{|V| \times |V|}$ be the weighted adjacency matrix of a network.  The {\em strength} $s(\cdot)$ of a node $i$ is given by the sum of the rows or columns of the matrix $W$,
\begin{equation*}
\label{eq:strength}
s(i) = \sum_{j=1}^{|V|}W_{ij} = \sum_{j=1}^{|V|}W_{ji}.
\end{equation*}
\end{definition}
Notice that the strength of a node is a weighted version of the degree of a node and therefore can be seen as a weighted, localized measure of node importance.

\begin{definition}
\label{def:clust_coeff}
Let $A = (A_{ij}) \in \{0,1\}^{|V| \times |V|}$ be the adjacency matrix of a network and $W$ its weighted version as above.  The {\em weighted clustering coefficient} (e.g., \cite{barrat2004architecture}) is given by
$$
c^{w}(i) = \frac{\sum_{j=1}^{|V|}\sum_{h=1}^{|V|}\{W_{ij} + W_{ih}\}A_{ij}A_{ih}A_{jh}}{2s(i)(d(i) - 1)}. 
$$
\end{definition}
While the standard clustering coefficient measures the proportion of possible connections between the neighbors of a given node, the weighted version takes into account the relative importance of the clustering structure around a given node by incorporating the edge weights.

%Additionally, we study the {\em weighted clustering coefficient} \cite{barrat2004architecture}, denoted by $c^w(i)$ in this work: while the standard clustering coefficient measures the proportion of possible connections between the neighbors of a given node, the weighted version takes into account the relative importance of the clustering structure around a given node by incorporating the edge weights.

%%%%%%%%%%%%%%%%%%%%%%%%%%%%%%%%%%%%%%%%%%%%%%%%%%%

\section{Implementation and Results}
\label{sec:results}

We implemented our proposed Forman-curved MCMC samplers given above in Sections \ref{sec:edgeMCMC} and \ref{sec:nodeMCMC} to approximate the network statistics given in Section \ref{sec:stats}.

Convergence performance for the curved samplers was compared to the base case of uniform sampling, where the transition to each neighbor node is with equal probability $p_{ij} = 1/d(i)$ for the edge-based approach and the uniform target distribution $\mathbb{P}(X_{k}= i) = {1}/{|V|}$ for the node-based approach.  Uniform sampling corresponds to standard MCMC approaches where curvature is not integrated into the method.

\subsection{Performance Evaluation}

To evaluate the performance of our proposed method, we compared the convergence of the network sample $G_n = \{V',E'\}$ to the full network $G$ by examining convergence to global network statistics across multiple Markov chains initialized at different starting points.

More specifically, for a given network statistic, let $Z$ denote its random variable for the full network.  Then for all network samples $n=1,2, \ldots, N$ and all chains $c=1,2, \ldots, n_c$, we examined the convergence of the estimator for the mean of each network statistic distribution
$$
\bar{Z}_{nc} = \frac{1}{|V'|}\sum_{j=1}^{|V'|} Z_{nc}(j)
$$
by calculating the mean squared error
$\displaystyle
\mathrm{MSE}_{n} = \frac{1}{n_c}\sum_{c=1}^{n_c}\{\bar{Z}_{nc}- \mathbb{E}[Z] \}^2.
$

\subsection{Application: Deterministic Real-World Networks}

We applied our curved edge-based and node-based MCMC samplers to deterministic networks arising from real-world data.  Specifically, we studied the character interaction network from the novel ``Les Mis\'{e}rables'' by Victor Hugo, with 77 nodes and maximum degree of 36 obtained from The KONECT Project \citep{konect} and the neural network of the {\em Caenorhabditis elegans} ({\em C. elegans}) worm with 306 nodes and maximum degree of 134 \citep{watts1998collective}.

\begin{figure}[ht]
  \centering
  \subfigure[\label{fig:LesMis_backbone_25}]{\includegraphics[width=0.45\columnwidth]{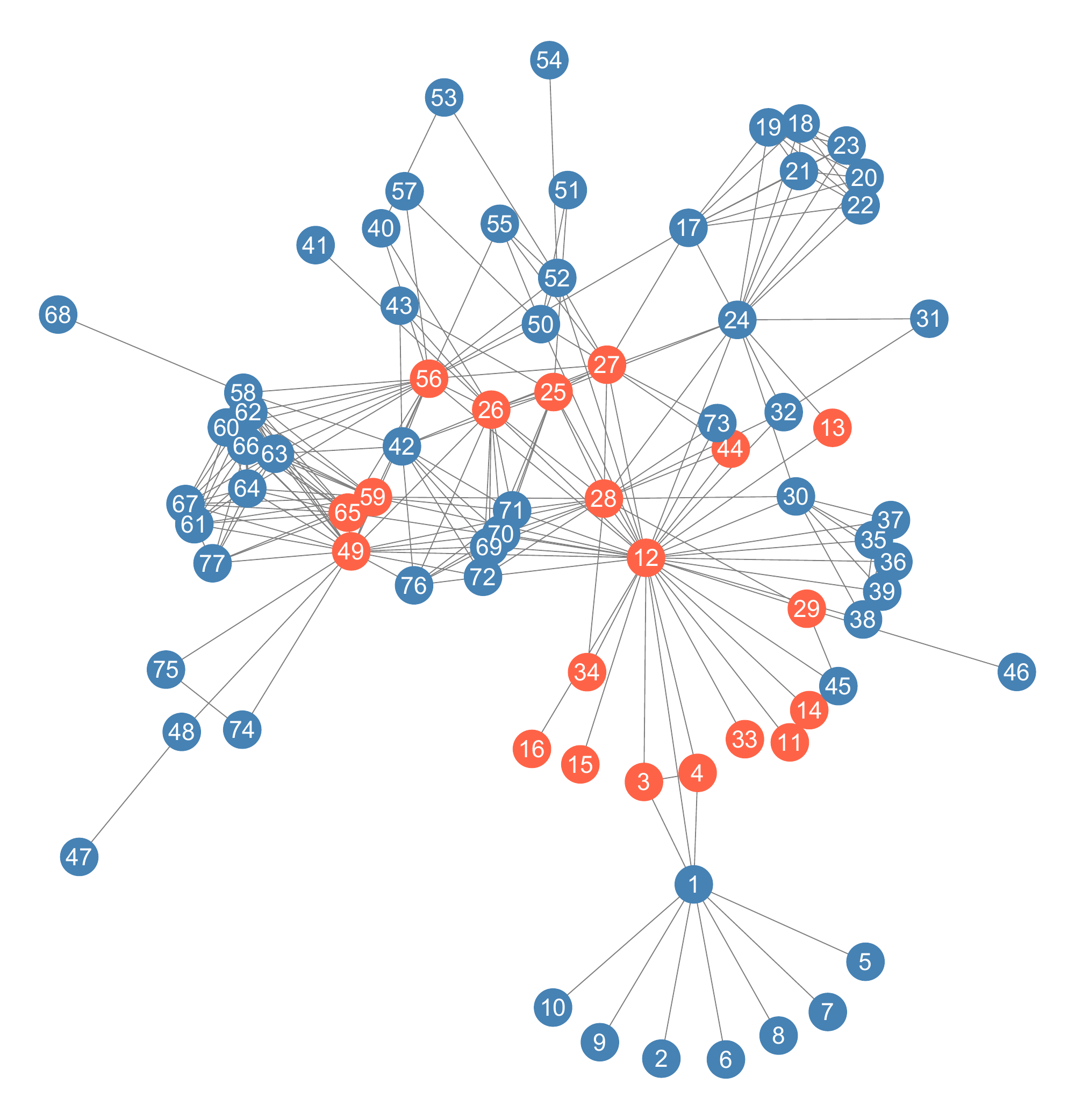}}
  \subfigure[\label{fig:LesMis_backbone_50}]{\includegraphics[width=0.45\columnwidth]{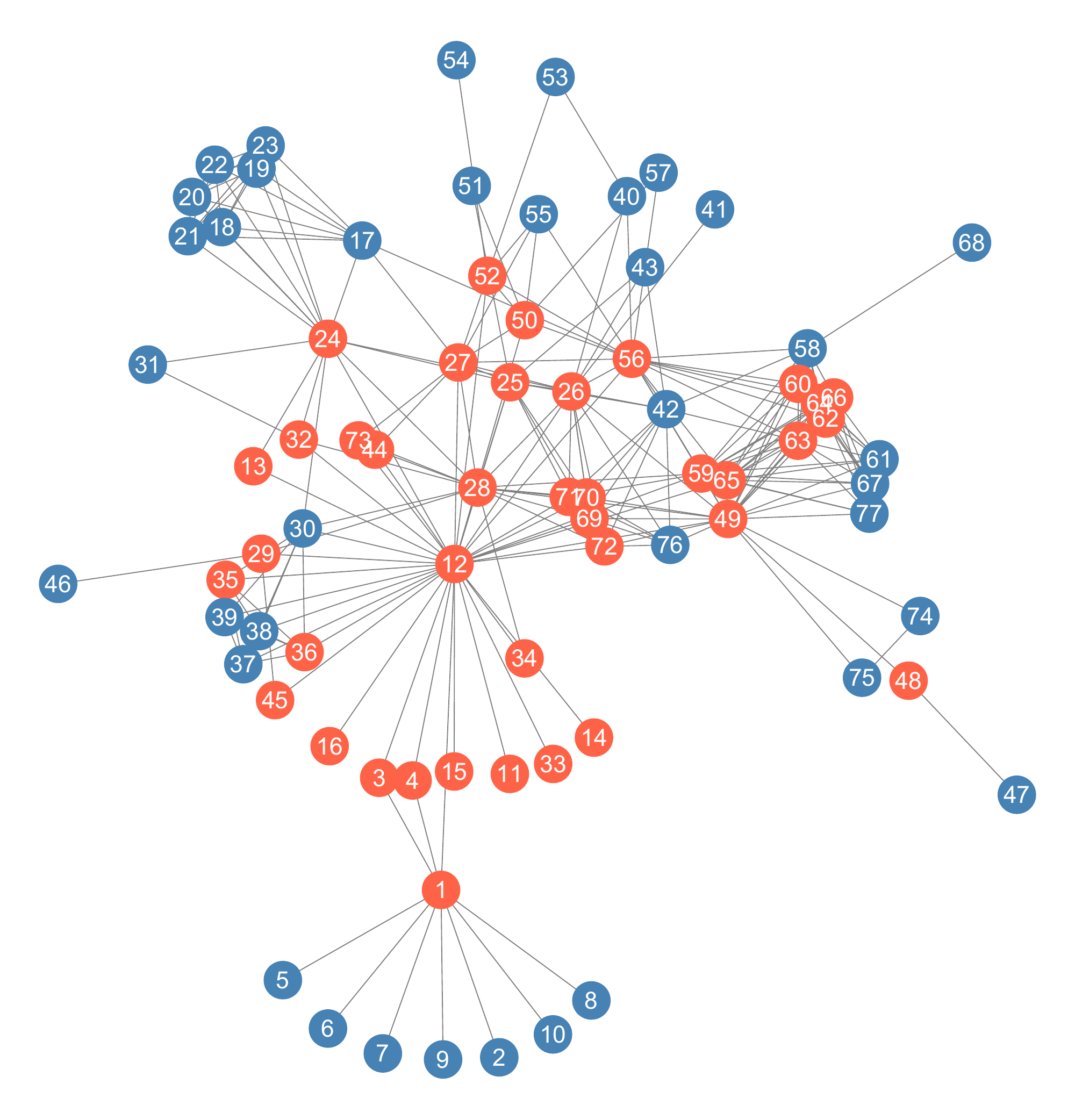}}
  \caption{\label{fig:LesMis_backbone} Les Mis\'{e}rables network: The blue nodes represent all nodes in the full network, while the orange nodes represent subsampled nodes by our proposed method.  (a) Top 25\% sampled nodes; (b) Top 50\% sampled nodes.}
\end{figure}

\begin{figure}[ht]
  \centering
  \subfigure[\label{fig:LesMis_edge_a}]{\includegraphics[width=0.45\columnwidth]{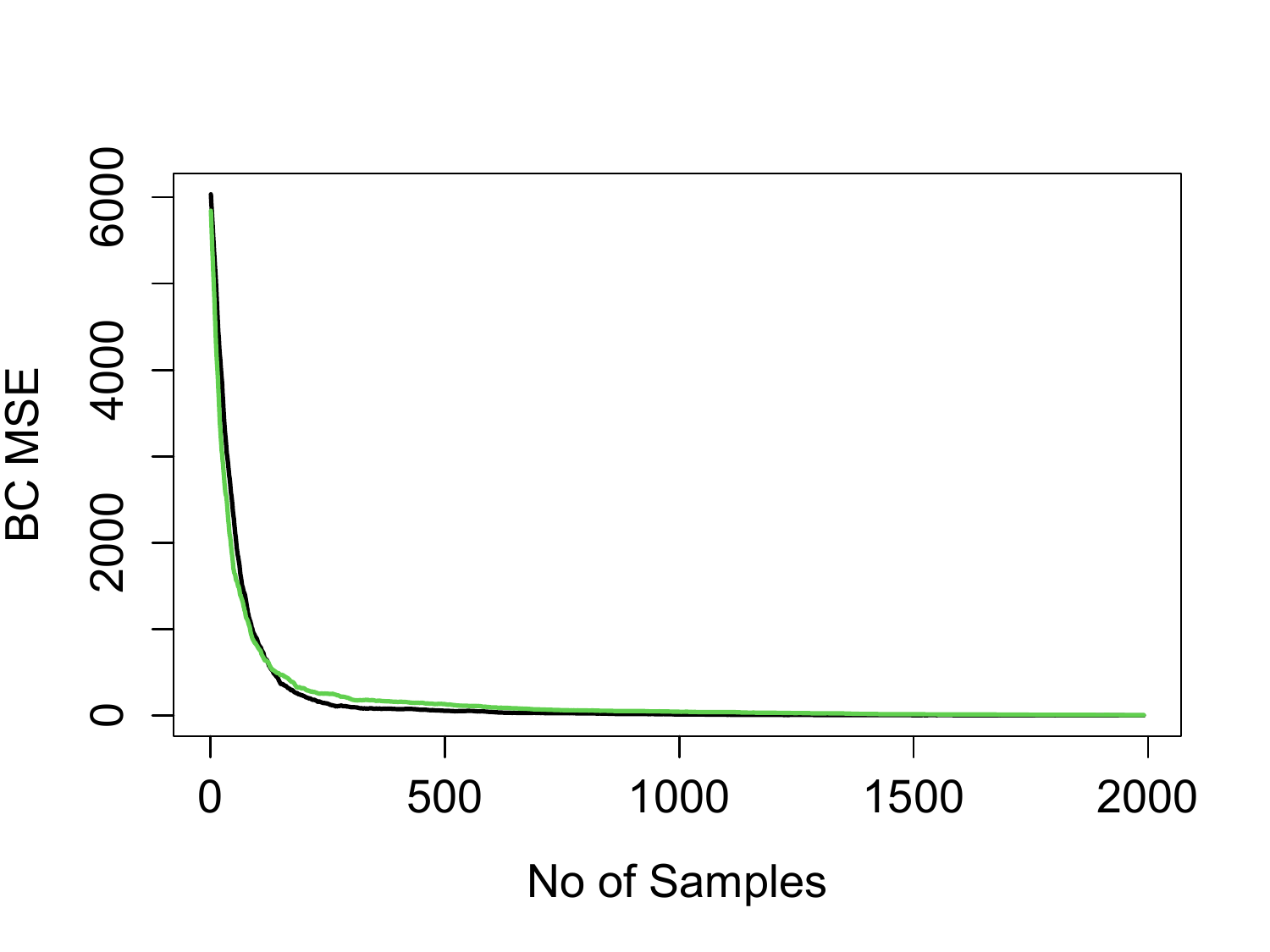}}
  \subfigure[\label{fig:LesMis_edge_b}]{\includegraphics[width=0.45\columnwidth]{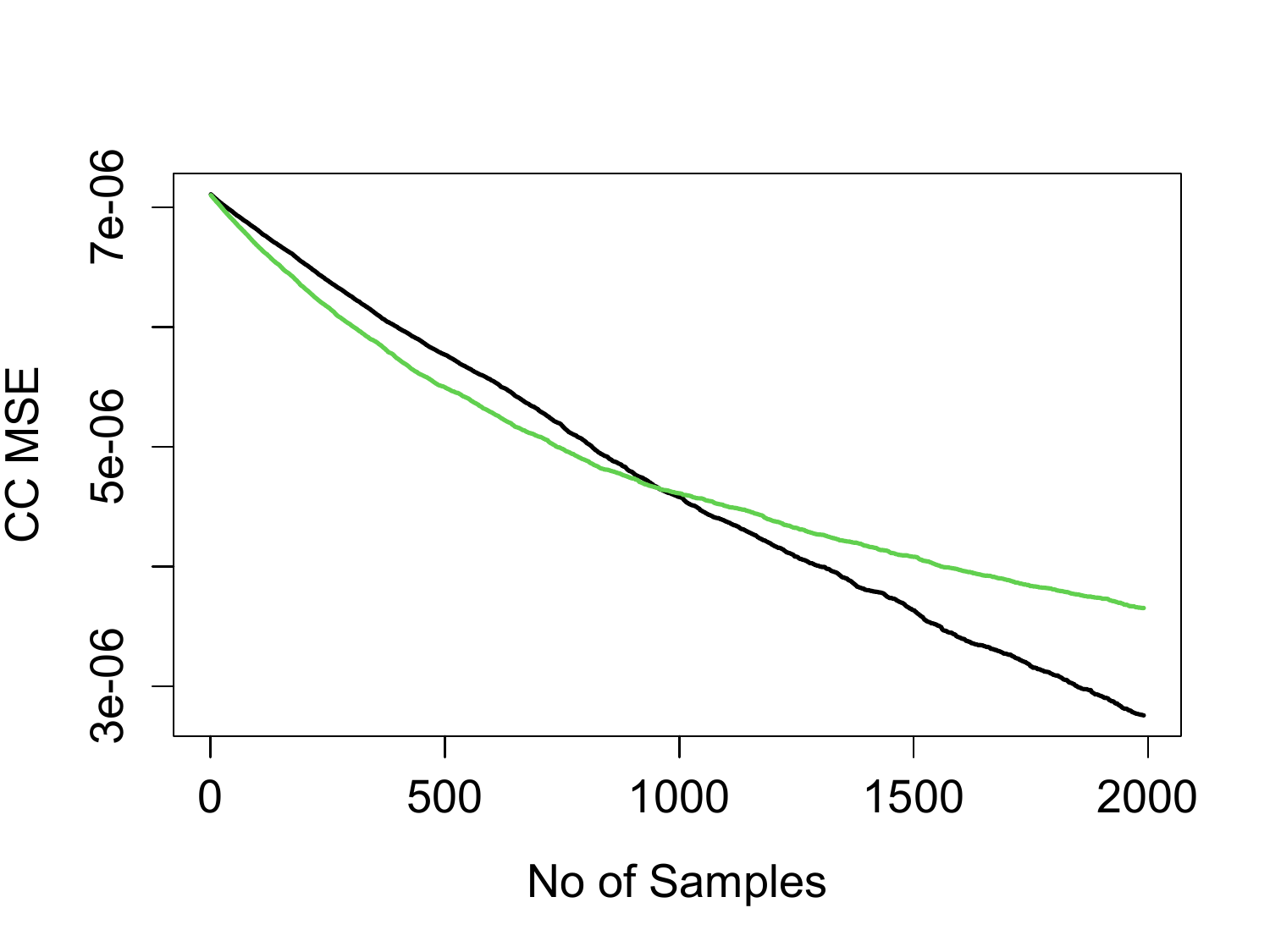}}
  \subfigure[\label{fig:LesMis_edge_c}]{\includegraphics[width=0.45\columnwidth]{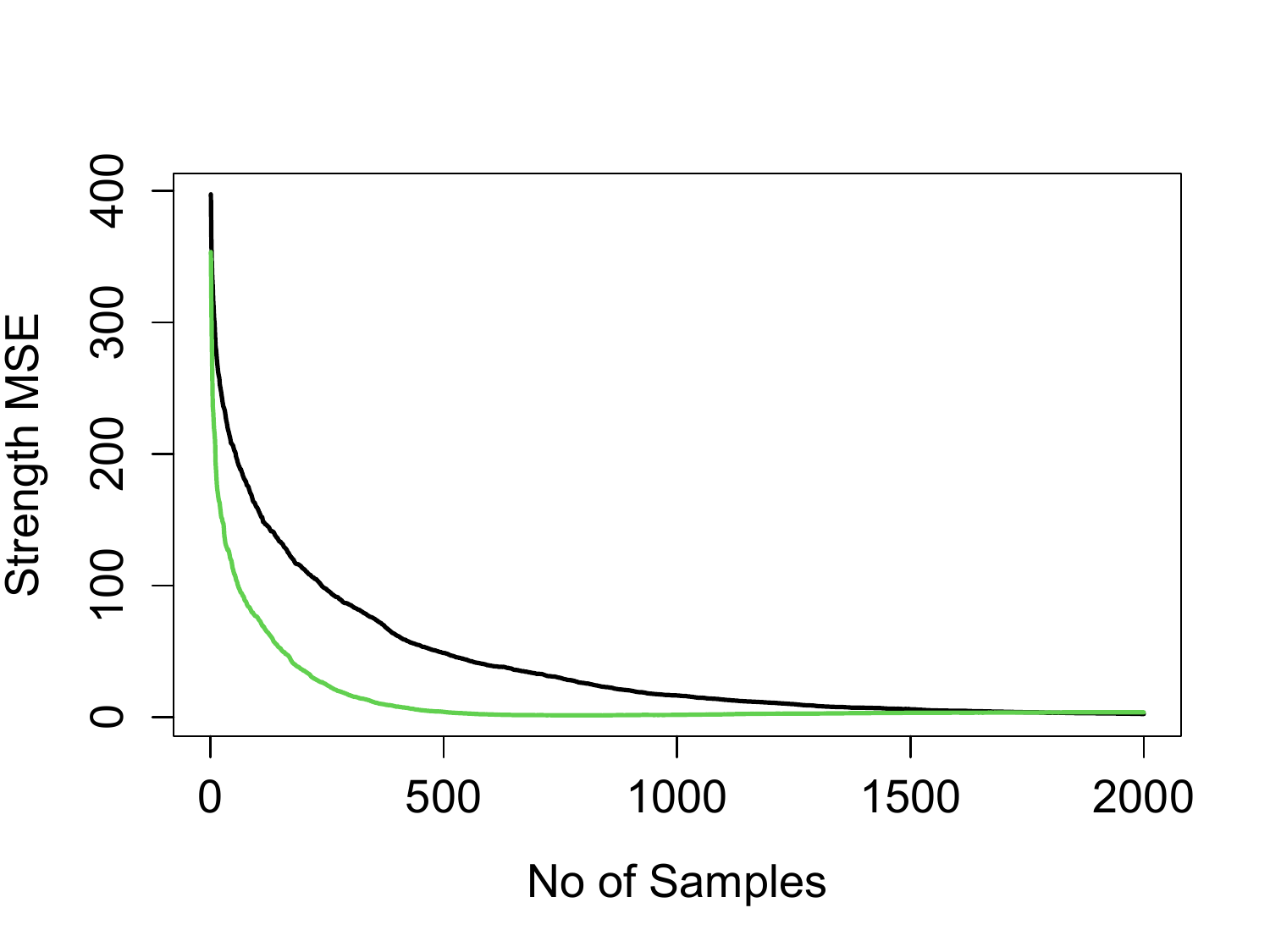}} 
  \subfigure[\label{fig:LesMis_edge_d}]{\includegraphics[width=0.45\columnwidth]{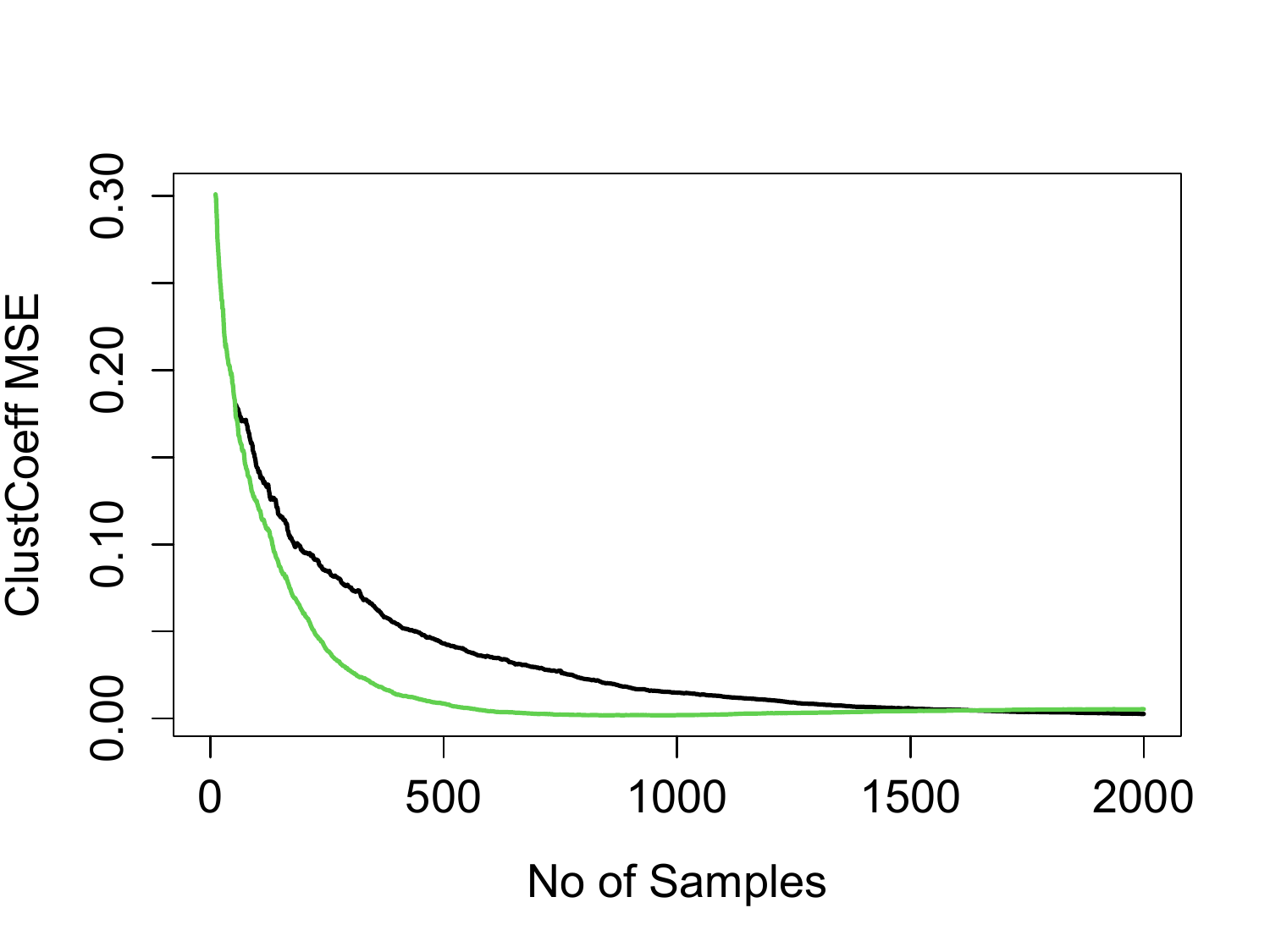}}
  \caption{\label{fig:LesMis_edge} Les Mis\'{e}rables network: Edge-based Forman-curved MCMC.  Black curve: Uniform sampling; Green curve: Sampling with transition probability (\ref{eq:mcmc_edge}). (a) Mean Betweenness Centrality; (b) Mean Closeness Centrality; (c) Mean Strength; (d) Mean Clustering Coefficient.}
\end{figure}

\begin{figure}[ht]
  \centering
  \subfigure[\label{fig:LesMis_node_a}]{\includegraphics[width=0.45\columnwidth]{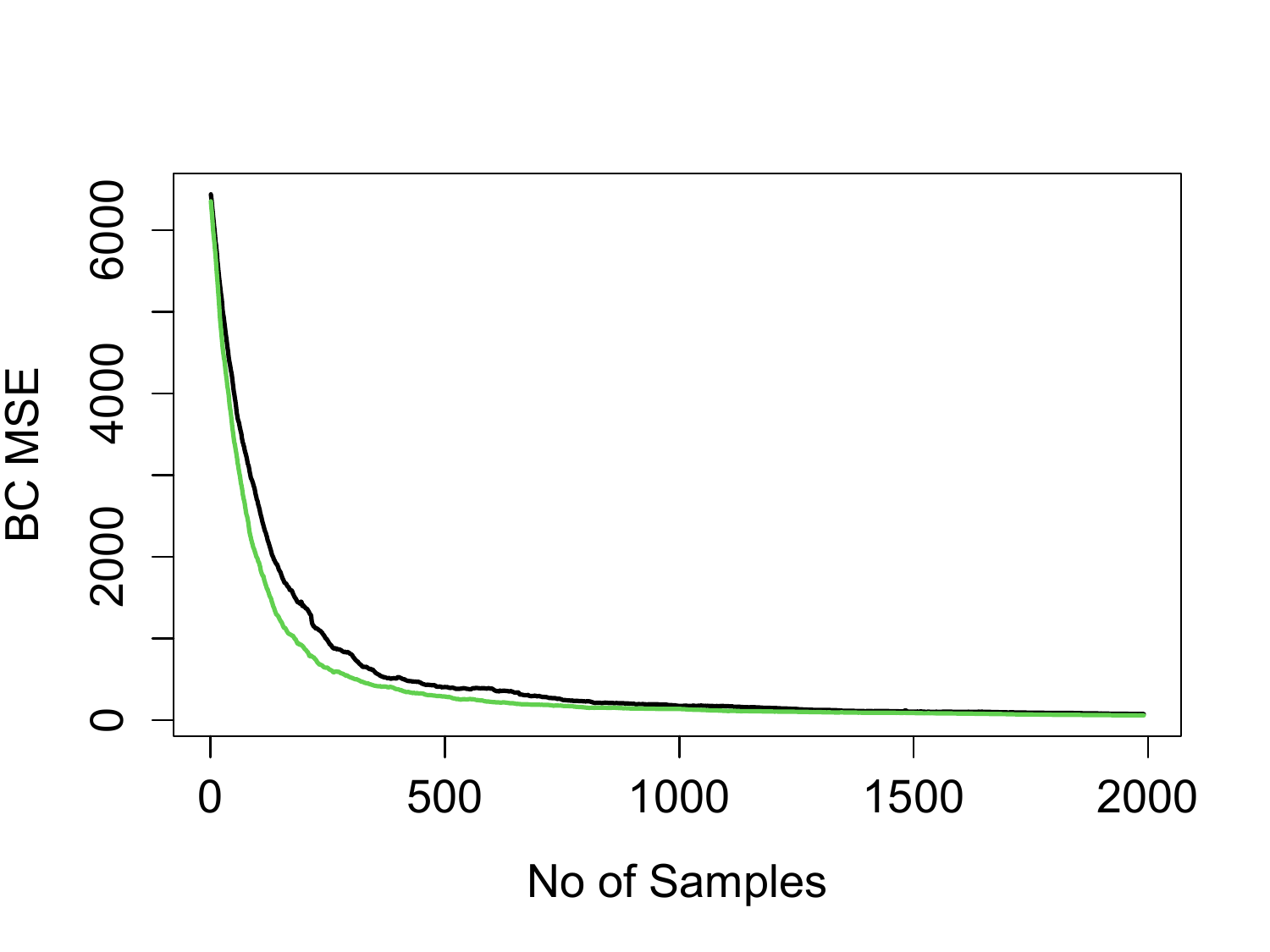}}
  \subfigure[\label{fig:LesMis_node_b}]{\includegraphics[width=0.45\columnwidth]{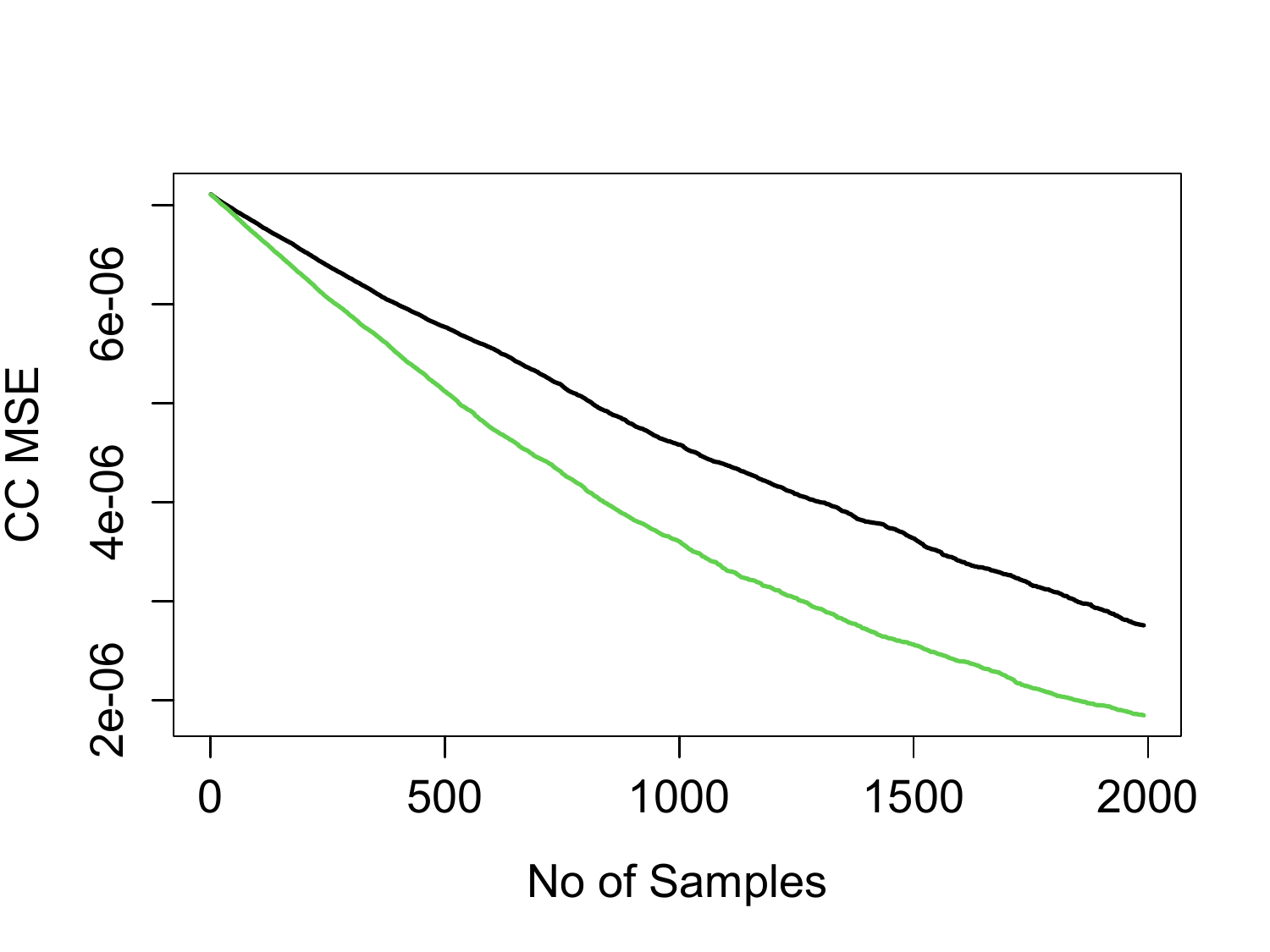}}
  \subfigure[\label{fig:LesMis_node_c}]{\includegraphics[width=0.45\columnwidth]{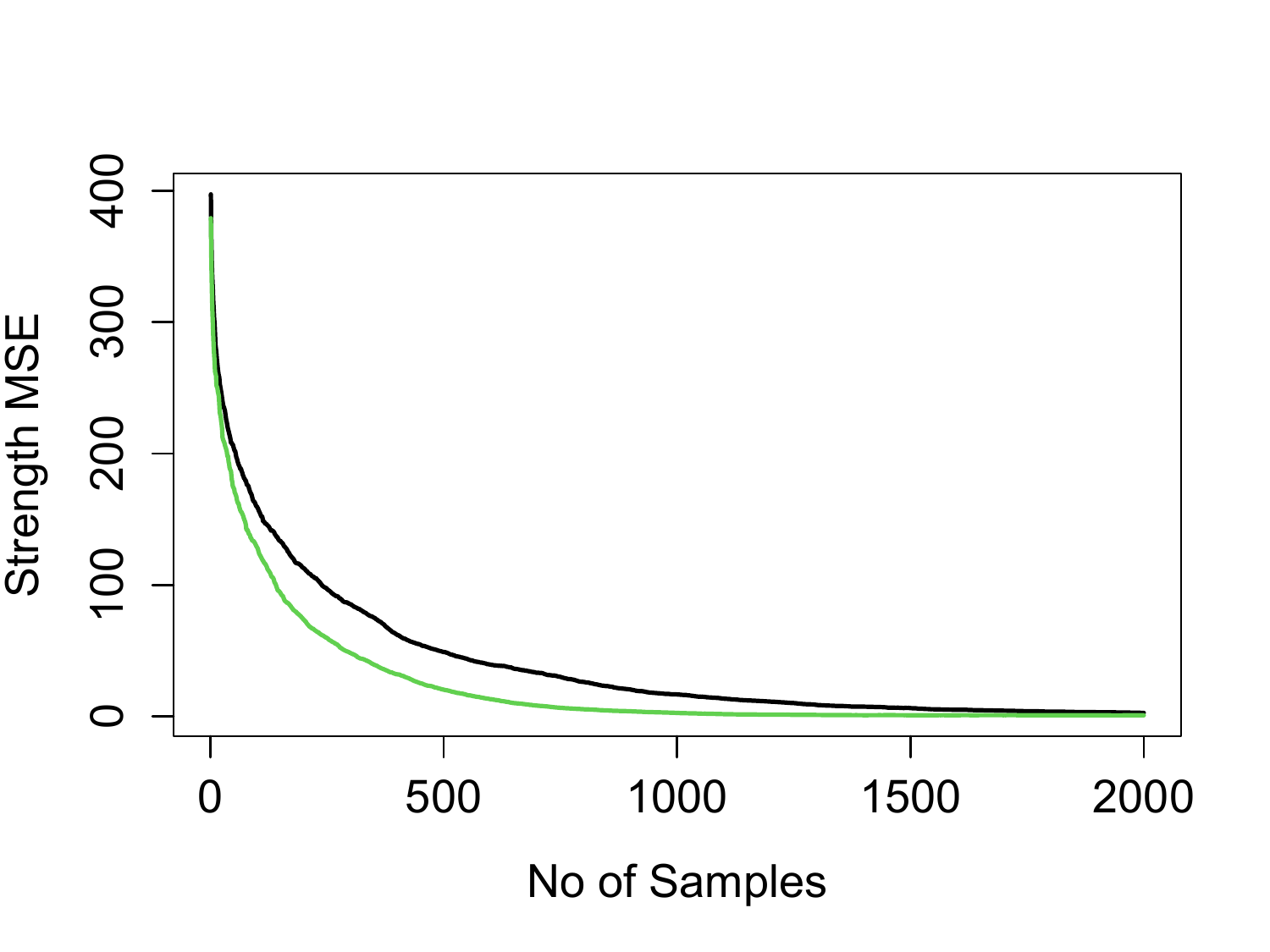}}
  \subfigure[\label{fig:LesMis_node_d}]{\includegraphics[width=0.45\columnwidth]{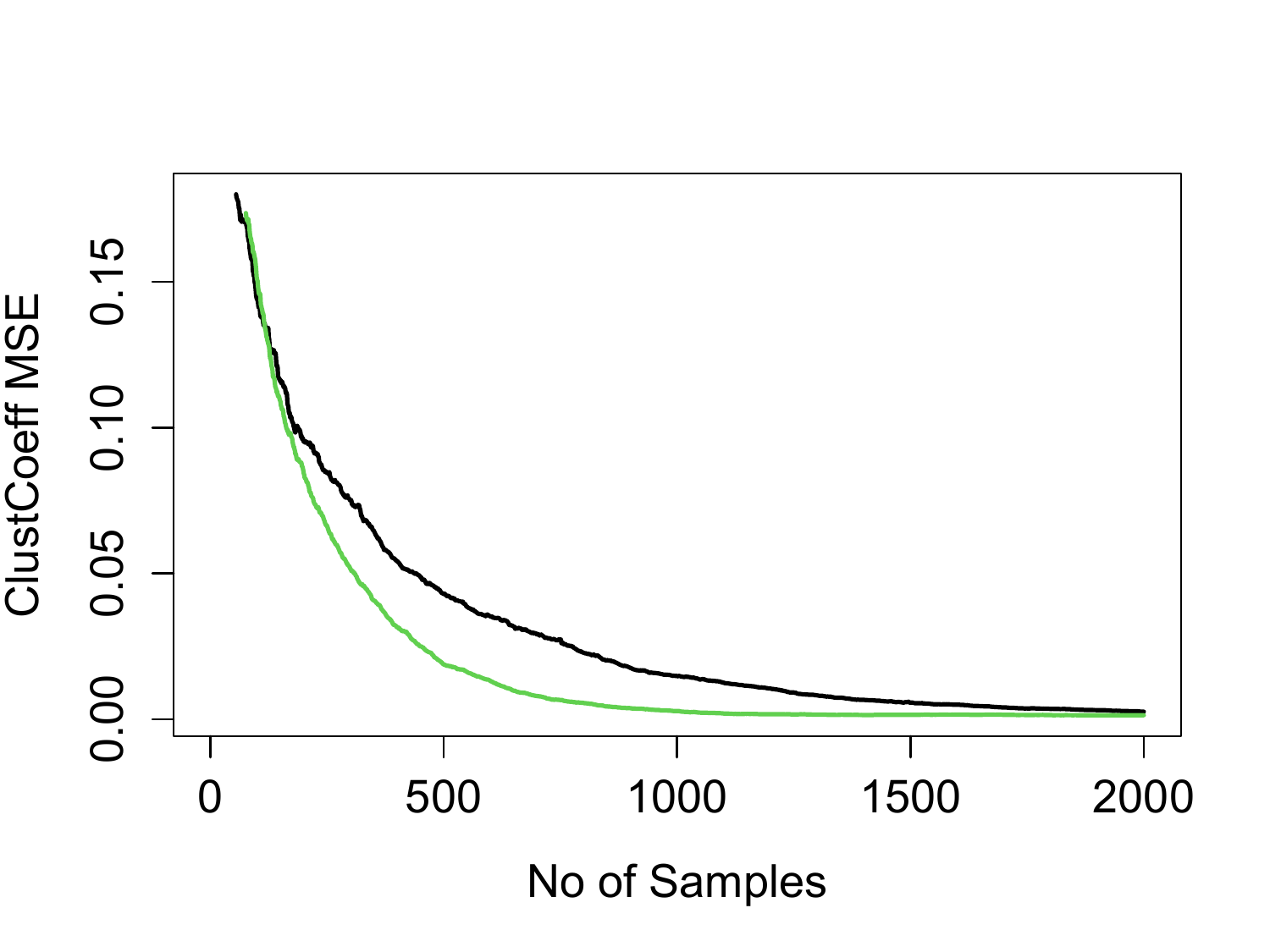}}
  \caption{\label{fig:LesMis_node} Les Mis\'{e}rables network: Node-based Forman-curved MCMC.  Black curve: Uniform sampling; Green curve: Sampling node $i$ with probability (\ref{eq:mcmc_node}). (a) Mean Betweenness Centrality; (b) Mean Closeness Centrality; (c) Mean Strength; (d) Mean Clustering Coefficient.}
\end{figure}

\begin{figure}[ht]
  \centering
  \subfigure[\label{fig:celegans_node_a}]{\includegraphics[width=0.45\columnwidth]{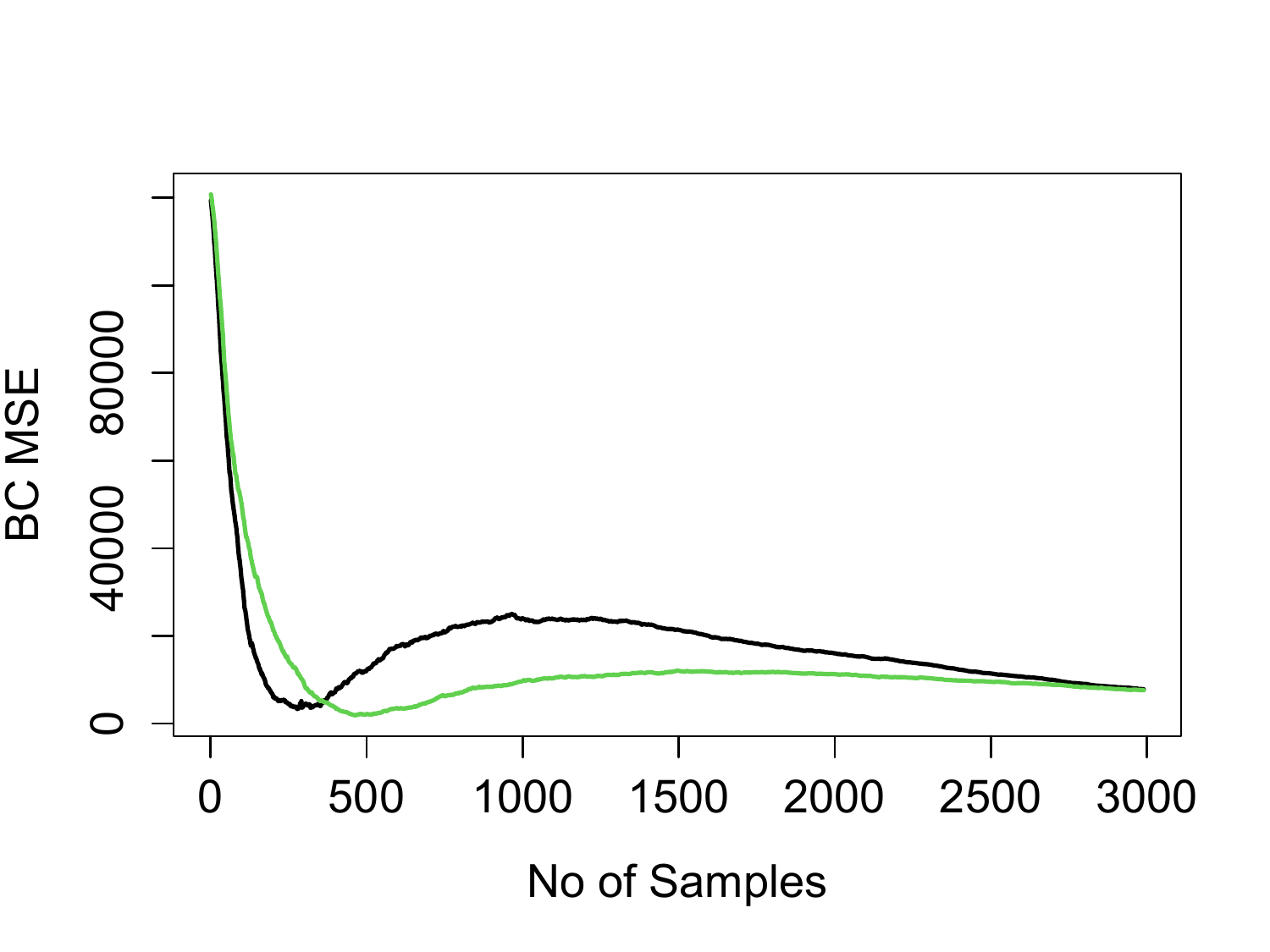}}
  \subfigure[\label{fig:celegans_node_b}]{\includegraphics[width=0.45\columnwidth]{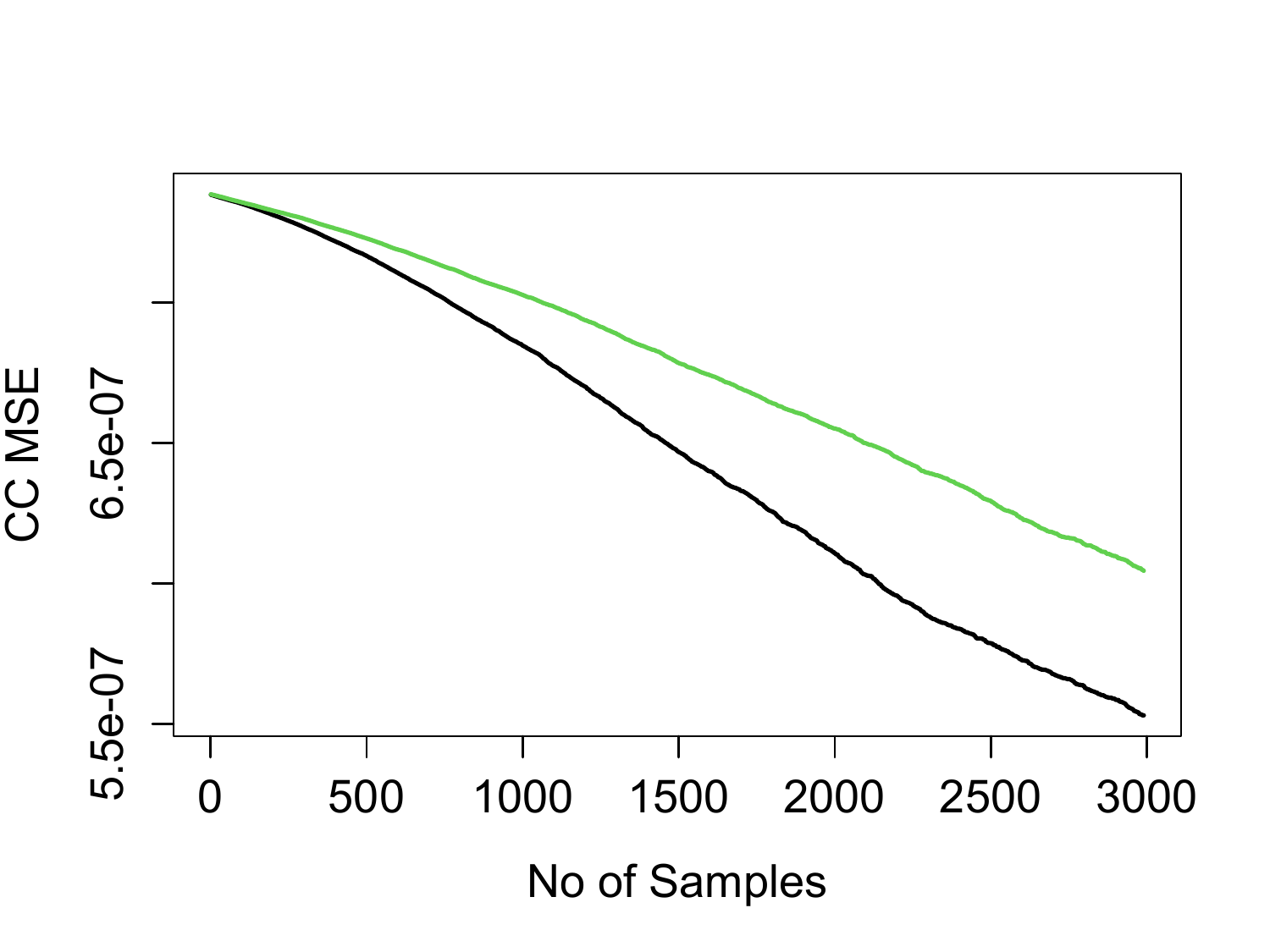}}
  \subfigure[\label{fig:celegans_node_c}]{\includegraphics[width=0.45\columnwidth]{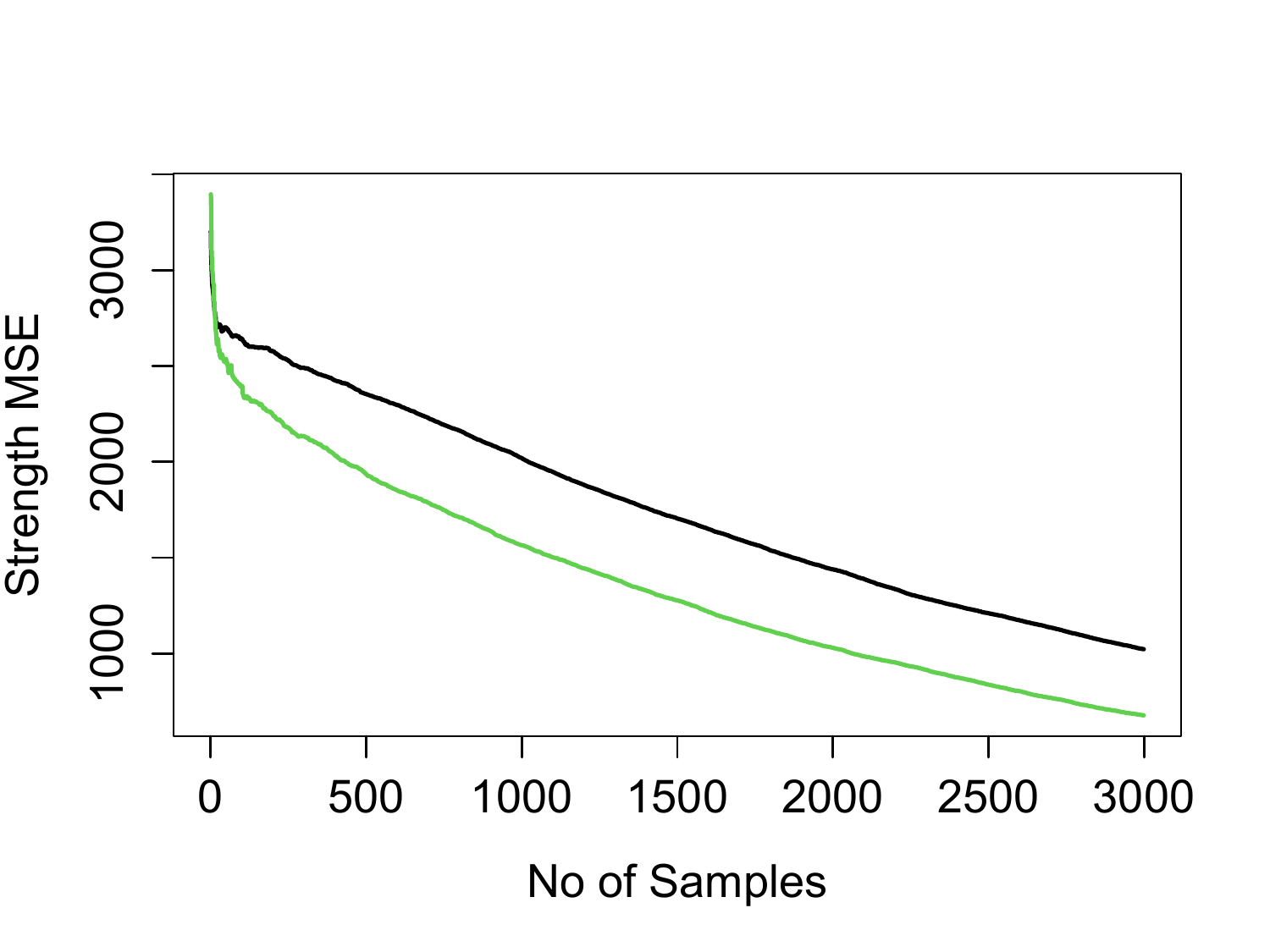}}
  \subfigure[\label{fig:celegans_node_d}]{\includegraphics[width=0.45\columnwidth]{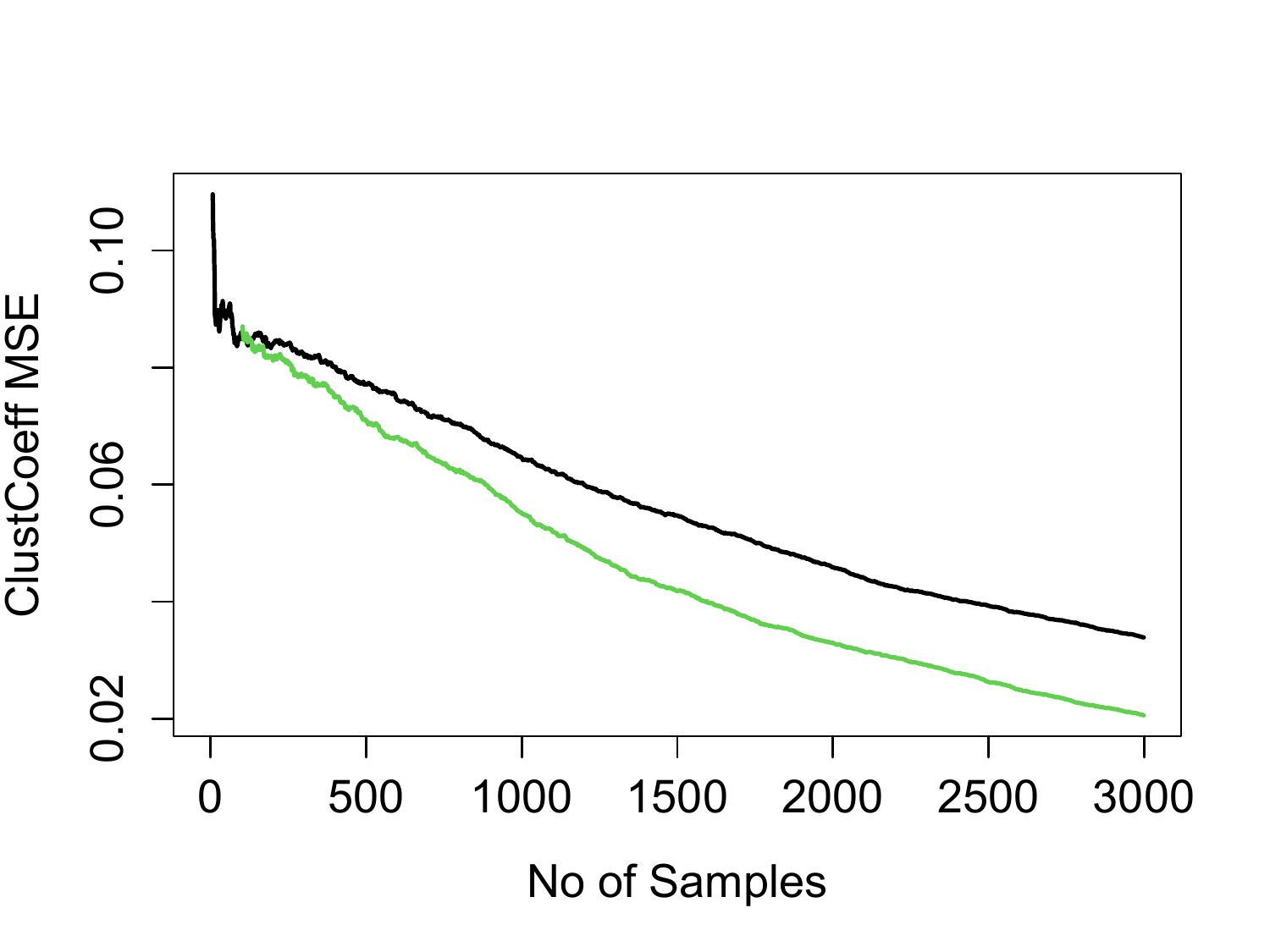}}
  \caption{\label{fig:celegans_node} {\em C. elegans} network: Node-based Forman-curved MCMC.  Black curve: Uniform sampling; Green curve: Sampling node $i$ with probability (\ref{eq:mcmc_node}). (a) Mean Betweenness Centrality; (b) Mean Closeness Centrality; (c) Mean Strength; (d) Mean Clustering Coefficient.}
\end{figure}

We implemented both the edge- and node-based curved Forman MCMC samplers proposed in Sections \ref{sec:edgeMCMC} and \ref{sec:nodeMCMC} to estimate all four network statistics listed in Section \ref{sec:stats} on the Les Mis\'{e}rables network; the results are displayed in Figures \ref{fig:LesMis_edge} and \ref{fig:LesMis_node}, respectively.  Examples of sampled subnetworks are displayed in Figure \ref{fig:LesMis_backbone}.  We implemented the node-based curved Forman MCMC sampler proposed in Section \ref{sec:nodeMCMC} to estimate all four network statistics listed in Section \ref{sec:stats} on the {\em C. elegans} network; the results are displayed in Figure \ref{fig:celegans_node}.

Across all sampling methods and networks, there is a visible advantage in integrating curvature into the MCMC sampling scheme.  In particular, the Forman-curved sampler outperforms the uniform sampler when estimating mean strength and clustering coefficient.  These results imply that the local behavior of graph Forman curvature is compatible with that of these network statistics.

For the two Forman-curved approaches applied to the Les Mis\'{e}rables network, we see similar convergence behavior to the uniform case for mean betweenness centrality.  For the edge-based curved sampler, we see an initial advantage in incorporating curvature which is then overtaken by the uniform sampler.  Overall, edge-based curved sampling converges faster than node-based curved sampling as shown in Figure \ref{fig:LesMis_edge} compared to Figure \ref{fig:LesMis_node}.  This is likely due to the uncorrelated node samples for the edge-based curved method and the fact that the methods view uniformity from two different levels.  Furthermore, it is in concordance with the fact that for networks, Ricci curvature is an edge-based measure.

There appears to be some evidence that mean betweenness centrality is better estimated by the curved Forman sampler for larger networks and for small sample sizes, which would benefit small sample reconstruction.  Further experiments on larger networks will need to be tested in order to corroborate this observation.

%Possible points to add:
%\begin{itemize}
%    \item Edge based sampling converges faster than node-based sampling (les mis edge vs les mis node). This is due to their being uncorrelated node samples for edge based method, and the fact that the methods view 'uniformity' from two different levels.
%    \item Across all networks, and sampling methods the Forman sampler outperforms the uniform sampler when estimating mean strength and clustering coefficient. This makes intuitive sense since Forman curvature is a local measure, and so are these network statisics.
%    \item Some evidence that BC is estimated better by Forman sampler for larger networks (c.elegans) for small sample sizes, which is beneficial for small sample reconstruction. However, more larger networks will need to be tested to buttress this.
%\end{itemize}

%%%%%%%%%%%%%%%%%%%%%%%%%%%%%%%%%%%%%%%%%%%%%%%%%%%

\section{Discussion}
\label{sec:discussion}

In this work, we studied the problem of network learning by sampling.  The impetus is that a larger network may be studied by an appropriate representation obtained by sampling; the retained network characteristics in the sample are determined by a wide variety of network statistics.  We estimated these statistics from the probabilistic viewpoint of Markov chain Monte Carlo and proposed a novel sampler that incorporates the discrete curvature of the network both explicitly and implicitly via both edge- and node-based approaches.  Specifically, curvature was integrated into the transition probability of the Markov chain as well as the target stationary distribution.  We found that our curved samplers noticeably improves the convergence to network statistics comprising local behavior. 

A particular advantage of our method is that graph Forman curvature is a local measure and thus requires only local information on the current node and its immediate neighbors.  In this sense, the technique can be viewed as semi-crawl-like and therefore is still applicable even without access to the full network.  This opens up possibilities for future research to develop semi-supervised network learning methodology by incorporating discrete curvature notions into reversible jump MCMC. 

Other natural directions of research include studying the accuracy of the estimated network statistics.  Other modifications to our samplers could also be studied; for instance, normalization by strength, which would result in bias towards nodes with largest weighted average, is a natural extension.  Additionally, it would be interesting to compare graph Forman--Ricci curvature with other types of Ricci curvature for networks as in \cite{barkanass2020geometric} and other geometrically motivated network measures.  Furthermore, additional experimentation on large-scale, real-world networks would help achieve a better understanding of the performance and applicability of our method on these scales, entailing further refinements of algorithms to increase their efficiency.

%%%%%%%%%%%%%%%%%%%%%%%%%%%%%%%%%%%%%%%%%%%%%%%%%%%

\section*{Acknowledgments}

We wish to thank Vladislav Barkanass for helpful discussions.  In addition, we wish to acknowledge the HPC facilities at the Department of Mathematics at Imperial College London utilized in this project. E.S.'s research is partially supported by the GIF Research Grant No.~I-1514-304.6/2019.

%%%%%%%%%%%%%%%%%%%%%%%%%%%%%%%%%%%%%%%%%%%%%%%%%%%

%\appendix
%\renewcommand{\thesection}{Appendix}
%\renewcommand{\thesubsection}{A\arabic{subsection}}
%\renewcommand{\theAppDefinition}{A\arabic{AppDefinition}}
%\renewcommand{\theAppClaim}{A\arabic{AppClaim}}
%\section*{Appendix}

%%%%%%%%%%%%%%%%%%%%%%%%%%%%%%%%%%%%%%%%%%%%%%%%%%%

\newpage
\section*{Figures}

%%%%%%%%%%%%%%%%%%%%%%%%%%%%%%%%%%%%%%%%%%%%%%%%%%%
%%%%%%%%%%%%%%%%%%%%%%%%%%%%%%%%%%%%%%%%%%%%%%%%%%%
%%%%%%%%%%%%%%%%%%%%%%%%%%%%%%%%%%%%%%%%%%%%%%%%%%%

% \section{Referencing}

% The bibliography file (a standard {\em .bib} formatted file) is % pulled in at the end, and articles, papers, books,
% web sites etc listed there are cited in various ways, such as \cite{carvalho2008bfrm} and \cite{aguilar2000jbes},
%as well as \citep{carvalho2008bfrm} and \citep{aguilar2000jbes}, and %also
%\citep{carvalho2008bfrm,aguilar2000jbes,west2003bayes7}. You can also use (\citet{aguilar2000jbes}) or
%(\citet{carvalho2008bfrm,aguilar2000jbes,west2003bayes7})  at a more hands-on level.

\clearpage
\newpage
\bibliography{curvedMCMC_ref} % edit bibexample.bib file ...
\bibliographystyle{chicago}  % or choose another bib list style

%%%%%%%%%%%%%%%%%%%%%%%%%%%%%%%%%%%%%%%%%%%%%%%%%%%
%%%%%%%%%%%%%%%%%%%%%%%%%%%%%%%%%%%%%%%%%%%%%%%%%%%
%%%%%%%%%%%%%%%%%%%%%%%%%%%%%%%%%%%%%%%%%%%%%%%%%%%

\end{document}